\documentclass[sigconf]{acmart}

\renewcommand\footnotetextcopyrightpermission[1]{} 
\makeatletter
\renewcommand\@formatdoi[1]{\ignorespaces}
\makeatother

\usepackage{booktabs} 
\usepackage{latexsym}
\usepackage{amsmath}
\usepackage{amsfonts}
\usepackage{url}
\usepackage{colortbl}
\usepackage{subcaption}
\usepackage{graphicx}
\usepackage{multicol}
\usepackage{numprint}
\npdecimalsign{.}%
\usepackage{siunitx}

\newcommand\n[1]{\numprint{#1}}

\usepackage{verbatim}
\usepackage{soul}

\newcommand\xdiag{\operatorname{diag}}
\newcommand\diag[1]{\xdiag\left(#1\right)}    

\newcommand\norm[1]{\left\lVert #1 \right\rVert}
\newcommand\vect[1]{{\boldsymbol{#1}}}

\newcommand\ve{\vect{e}}

\newcommand\vr{\vect{r}}

\newcommand\mR{\vect{R}}

\newcommand\bN{\mathbb{N}} 

\newcommand\bR{\mathbb{R}} 

\DeclareMathAlphabet{\mathcal}{OMS}{cmsy}{m}{n}

\newcommand\cE{\mathcal{E}}

\newcommand\cK{\mathcal{K}}

\newcommand\cR{\mathcal{R}}

\newcommand\cT{\mathcal{T}}

\DeclareMathAlphabet\mathbfcal{OMS}{cmsy}{b}{n}

\acmYear{2019}
\copyrightyear{2019}

\begin{document}

\title{On Evaluating Embedding Models for Knowledge Base Completion}

\author{Yanjie Wang}
\affiliation{%
  \institution{University of Mannheim}
  \city{Mannheim}
  \state{Germany}
}
\email{ywang@uni-mannheim.de}

\author{Daniel Ruffinelli}
\affiliation{%
  \institution{University of Mannheim}
  \city{Mannheim}
  \state{Germany}
}
\email{daniel@informatik.uni-mannheim.de}

\author{Rainer Gemulla}
\affiliation{%
  \institution{University of Mannheim}
  \city{Mannheim}
  \state{Germany}
}
\email{rgemulla@uni-mannheim.de}

\author{Samuel Broscheit}
\affiliation{%
  \institution{University of Mannheim}
  \city{Mannheim}
  \state{Germany}
}
\email{broscheit@informatik.uni-mannheim.de}

\author{Christian Meilicke}
\affiliation{%
  \institution{University of Mannheim}
  \city{Mannheim}
  \state{Germany}
}
\email{christian@informatik.uni-mannheim.de}

\begin{abstract}
  Knowledge bases contribute to many web search and  mining tasks, yet they 
  are often incomplete. To add missing facts to a given knowledge base, 
  various embedding models have been proposed in the recent literature. 
  Perhaps surprisingly, relatively simple models with limited expressiveness 
  often performed remarkably well under today's most commonly used evaluation 
  protocols. In this paper, we explore whether recent models work well for 
  knowledge base completion and argue that the current evaluation protocols 
  are more suited for question answering rather than knowledge base completion. 
  We show that when focusing on a different prediction task for evaluating
  knowledge base completion, the performance of current embedding models is 
  unsatisfactory even on datasets previously thought to be too easy. This is 
  especially true when embedding models are compared against a simple 
  rule-based baseline.
  This work indicates the need for more research into the embedding models and 
  evaluation protocols for knowledge base completion.
\end{abstract}


\maketitle

\section{Introduction}

Knowledge bases (KB) such as
DBpedia~\cite{AuerBKLCI07}
or YAGO~\cite{DBLP:conf/semweb/RebeleSHBKW16}
have become valuable resources for many web applications, including web search~\cite{Search,Search2}, 
information extraction~\cite{Events}, and question
answering~\cite{QA2,DBLP:conf/emnlp/GuML15,DBLP:conf/acl/XuRFHZ16}, 
A knowledge base is a collection of relational facts, often represented in form of 
\emph{(subject, relation, object)}-triples; e.g., \emph{(Einstein, bornIn, Ulm)}. 
Despite the great effort in constructing large KBs, they still miss a large number of
facts~\cite{DBLP:conf/www/WestGMSGL14}. Consequently, there has been
considerable interest in the task of knowledge base completion (KBC), which aims
to automatically infer missing facts by reasoning about the information already
present in the KB.

In the recent literature, a large number of \emph{embedding models} for KBC have
been proposed. Such models~\cite{nickel2016review} aim at embedding both entities 
and relations in a low-dimensional latent factor space such that the structure of
the knowledge graph is suitably captured. The embeddings are subsequently used
to score unobserved triples in order to assess whether they constitute missing
information (high score) or are likely to be false (low score).
\citet{DBLP:conf/icml/LiuWY17} and \citet{DBLP:journals/corr/abs-1709-04808}
recently showed that some embedding models are restricted in expressiveness in
the sense that they cannot model all types of relations, most notably
TransE~\cite{DBLP:conf/nips/BordesUGWY13} and
DistMult~\cite{DBLP:journals/corr/YangYHGD14a}. Nevertheless, these models
showed competitive empirical results in multiple studies. Moreover, it was
repeatedly shown that simple baselines outperformed the best embedding models
for KBC, raising concerns about the benchmarks~\cite{toutanova2015observed}, the
models~\cite{DBLP:conf/rep4nlp/KadlecBK17} and the
evaluation~\cite{DBLP:journals/corr/abs-1710-10881}. 

In this paper, we argue that the methods used to evaluate these models are more
suitable for question answering (QA) than for KBC. The most commonly
used evaluation protocol\footnote{We discuss other less adopted evaluation methods in
  Sec.~\ref{sec:discussion}.}  for KB embedding models is the
\emph{entity ranking protocol} (ER), which uses held-out test data to assess
model performance. In particular, for each true test triple such as
\emph{(Einstein, bornIn, Ulm)}, the embedding model is used to rank answers to
the questions \emph{(?, bornIn, Ulm)} and \emph{(Einstein, bornIn, ?)}. Model
performance is then assessed based on the rank of the test triple in the result.
ER thus assesses a model's performance for KBC based on its ability to answer
certain questions. The evaluation set is constructed such that these questions
are ``sensical'' and they always have an answer. As a result, models that assign
high scores to nonsensical triples such as \emph{(Ulm, bornIn, Einstein)} are
not penalized because the corresponding questions are never asked.

Unlike QA, where the focus is on answering meaningful questions, KBC has
different goals: (1) to add missing true triples to the knowledge base and (2)
to avoid adding false triples. The first point relates to recall, while the
second relates to precision. In fact, a model that performs well under ER may
have low precision overall, i.e., it may assign high scores to a large number of
false triples (e.g., nonsensical triples). If we completed a KB using the
high-scoring triples, we would add these false triples and deteriorate
precision. This is undesirable, since many KBs are constructed to be highly
precise---e.g.\ YAGO has a precision of 95\%~\cite{suchanek2008yago}---and a drop
in precision would negatively affect downstream applications. 
Thus,  models that assign high scores to false triples should be penalized.


In this paper, we propose that instead of asking $(i, k, ?)$ and $(?, k, j)$, 
a more appropriate question for KBC is $(?, k, ?)$, e.g. \emph{(?, bornIn, ?)},
since in KBC we are interested in any missing facts for every relation $k$.
To estimate the performance of the models when answering such questions, 
we explore a new protocol for automatic evaluation called \emph{entity-pair ranking} (PR). 
For a given relation $k$, PR ranks answers to the question $(?, k, ?)$, e.g.\ 
\emph{(?, bornIn, ?)}, meaning that the rank of all test triples of relation 
\emph{bornIn} are compared against all possible
answers, i.e., entity pairs in the knowledge graph. This includes nonsensical
triples such as \emph{(Ulm, bornIn, Einstein)}. Consequently, model performance
when evaluating with PR is negatively affected when models assign high scores to
false or even nonsensical triples.

The central research question in this work is: do embedding models work well for KBC? 
To that end, we conducted an extensive set of experiments to evaluate the performance of
well-known embedding models on multiple datasets using ER and PR. 
When comparing embedding models with a simple rule-based approach on standard datasets, 
we found that PR generally shows the performance of all models to be unsatisfactory for KBC,
which is unlike what is suggested by ER. 
Additionally, we analyzed the underestimation effect present in all evaluation
protocols due to unobserved true triples predicted by the models. This issue is inherent in KBC 
and it is generally unavoidable for automatic methods, i.e. without human labeling. We found that
while PR suffers from this effect as well, it can still be useful to assess model
performance for KBC. Finally, to see whether the performance
of embedding models was mostly due to the nonsensical triples considered in PR, 
we applied the background knowledge of domain-range constraints to filter out many 
of these clearly nonsensical triples. We found that embedding models indeed suffer 
from nonsensical triples, and that their performance is still unsatisfactory even
after removing these triples during evaluation. 


Our results indicate that entity ranking is not suitable for KBC, as it considerably overestimates 
the performance of embedding models. In addition, we observed that when compared against a simple 
rule-based baseline, current embedding models do not perform well even on ``simple'' datasets. 
This implies the need for more powerful models. 
To the best of our knowledge, this is the first work which elaborates on the
problems of current evaluation protocols, and shows that current embedding models
have unsatisfactory performance even on the benchmark datasets which were thought to be too easy
for embedding models.

This paper is organized as follows: in Section~\ref{sec:preliminaries} we introduce
the basic concepts of embedding-based, as well as rule-based models for KBC. In
Section~\ref{sec:eval} we describe the currently used evaluation protocols for KBC, 
discuss their shortcomings, and propose an alternative protocol for the experimental study 
in Section~\ref{sec:experiments}. We draw conclusions in Section~\ref{sec:conclusion}.


\section{Preliminaries}\label{sec:preliminaries}

Given a set of entities $\cE$ and a set of relations $\cR$, a knowledge base
$\cK\subseteq \cE\times\cR\times\cE$ is a set of triples $(i, k, j)$, where 
$i, j \in \cE$ and $k \in \cR$. Commonly, $i, k$ and $j$ are referred to as 
the \emph{subject}, \emph{relation}, and \emph{object}, respectively. A knowledge 
base can be viewed as a labelled graph, where each vertex corresponds to an entity, 
each label to a relation, and each labeled edge to a triple.

An \textit{embedding model} associates an \textit{embedding} $\ve_i\in\bR^{d}$
and $\vr_k\in\bR^{d_R}$ in a low-dimensional vector space with each entity $i$ and
relation $k$, respectively. We refer to hyperparameters $d,d_R\in\bN^+$ as the
\emph{size} of the embeddings. Each model uses a \textit{scoring function}
$s:\cE \times \cR \times \cE \rightarrow \mathbb{R}$ to associate a score $s(i,k,j)$ 
to each triple $(i,k,j)\in \cE\times\cR\times\cE$. The scores induce a ranking: 
triples with high scores are considered more likely to be true than triples with 
low scores. The scoring function depends on $i$, $k$, and $j$ only through their 
respective embeddings $\ve_i$, $\vr_k$, and $\ve_j$. Roughly speaking, the models 
try to find embeddings that capture the structure of the entire knowledge graph well. 
Since embeddings constitute a form of compression, the models are forced to
generalize so that new facts can be predicted.

In the following, we briefly review some recent embedding models for knowledge
base completion. We focus throughout on the set of models investigated
with respect to their expressiveness
in~\citet{DBLP:journals/corr/abs-1709-04808,DBLP:conf/icml/LiuWY17}. This allows
us to compare model expressiveness with empirical model performance. Many more
models have been proposed in the literature; e.g. the neural models
GCN~\cite{DBLP:journals/corr/SchlichtkrullKB17} and
ConvE~\cite{dettmers2018conve}. 

\paragraph{Embedding models.}
We subsequently write $\mR_k\in\bR^{d\times d}$ (instead of $\vr_k$) for the 
embedding of relation $k$ if it is best interpreted as a matrix. Then $d_R=d^2$;
otherwise, $d_R=d$.

RESCAL~\cite{DBLP:conf/icml/NickelTK11} is the most general bilinear model and uses 
the scoring function $s(i,k,j) = \ve_i^T\mR_k\ve_j$.
TransE~\cite{DBLP:conf/nips/BordesUGWY13} is a translation-based model inspired
by Word2Vec~\cite{DBLP:journals/corr/abs-1301-3781} and uses the score function $ s(i,k,j) =
-\norm{\ve_i+\vr_k - \ve_j}$ (using either $l_1$ or $l_2$ norm).
DistMult~\cite{DBLP:journals/corr/YangYHGD14a,Carroll1970} is a fairly constrained
factorization model with scoring function
$ s(i,k,j) = \ve_i^T\diag{\vr_k}\ve_j$.
ComplEx~\cite{DBLP:conf/icml/TrouillonWRGB16} uses embeddings in the complex
domain with scoring function $s(i,k,j) = \operatorname{Real}(\ve_i^T\diag{\vr_k}\ve_j)$, 
where $\operatorname{Real}(\cdot)$ extracts the real part of a complex number.
Analogy~\cite{DBLP:conf/icml/LiuWY17} uses the scoring function $
s(i,k,j)=\ve_i^T\mR_k\ve_j $ of RESCAL, but constrains $\mR_k\in \mathbb{R}^{d\times
  d}$ to a block diagonal matrix in which each block is either a real scalar or
a $2\times2$ matrix of form $\begin{pmatrix}
  x & -y \\
  y & x
  \end{pmatrix}$ with $x,y\in\bR$.

Unlike RESCAL, Analogy and ComplEx, DistMult and TransE are restricted models in that
they cannot represent any given knowledge base. See Sec.~\ref{sec:discussion} for more details.

\paragraph{Rule learning.}
A traditional approach to relational tasks is rule learning~\cite{Getoor}. Recent examples of
successful rule learning systems are AMIE and RuleN~\cite{amie2013,meilicke2018fine}. Generally 
speaking, such an approach learns rules which encode dependencies found in the KB. The types of
rules a given system can learn are known as its \emph{language bias}. Specific rules are learned by
looking at groundings of these possible rules in the data. These instances will determine the rule's 
confidence. Subsequently, the learned probabilistic rules are used in combination with the KB
to predict missing facts based on their assigned scores.


\section{Evaluation Protocols}\label{sec:eval}

Previous work questioned the effectiveness of 
benchmark datasets~\cite{toutanova2015observed,dettmers2018conve}.
In contrast, in this section we point out the issues with the current evaluation
protocols for KBC. To this end, we first review two widely used evaluation protocols. 
We then argue that these protocols are not well-suited for assessing KBC performance,
because they focus on a small subset of all possible facts for a given relation,
and thus the overall precision of the models is not reflected in the results.
To illustrate this point, we describe the PR protocol and discuss its advantages and
potential shortcomings.

\subsection{Current Evaluation Protocols}

Most studies use the triple classification (TC) and the entity ranking (ER) 
protocols to assess model performance, where ER is arguably the most 
widely adopted protocol. We assume throughout that only true but no false triples 
are available (as is commonly the case), and that these triples are divided
into training, validation, and test triples. The union of these three sets acts as
a proxy of the entire KB, which is unknown due to incompleteness.

\paragraph{Triple classification (TC)} The goal of TC is to test
the model's ability to discriminate between true and false
triples~\cite{DBLP:conf/nips/SocherCMN13}. Since only true triples are available
in practice, pseudo-negative triples are generated by randomly replacing either
the subject or the object of each test triple by another random entity that appears
as a subject or object, respectively. Each
resulting triple is then classified as positive or negative. In particular,
triple $(i,k,j)$ is classified as positive if its score $s(i,k,j)$ exceeds a
relation-specific decision threshold $\sigma_k$ (learned on validation data using 
the same procedure). Model performance is assessed by its accuracy, i.e., 
how many triples are classified correctly.

\begin{figure}
  \centering
  \begin{subfigure}{0.35\textwidth}
    \centering
    \includegraphics[scale=0.32]{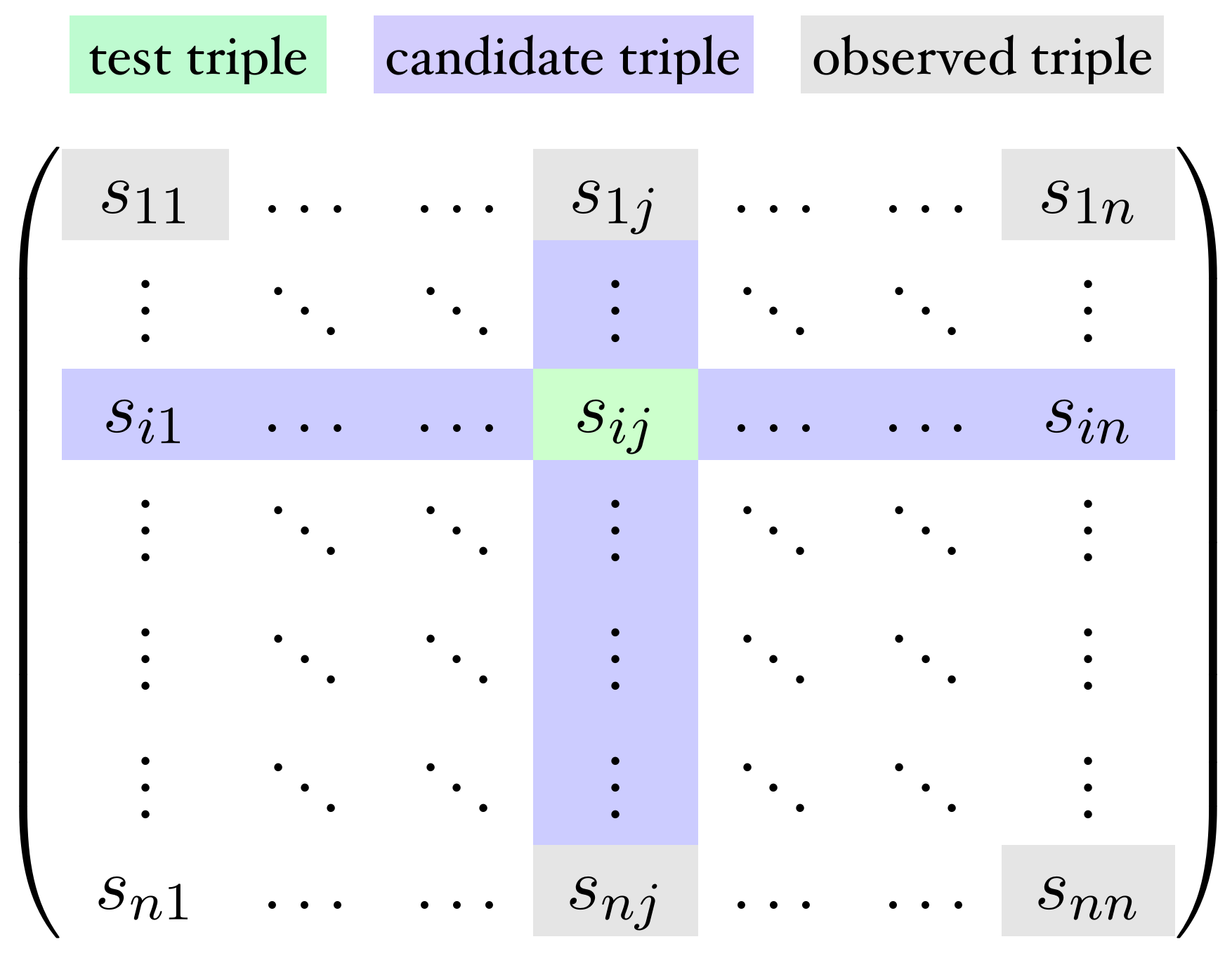}
    \caption{Entity Ranking protocol (ER)}
    \label{fq:protocol}
  \end{subfigure}
\\\par
\begin{subfigure}{0.35\textwidth}
  \centering
    \includegraphics[scale=0.4]{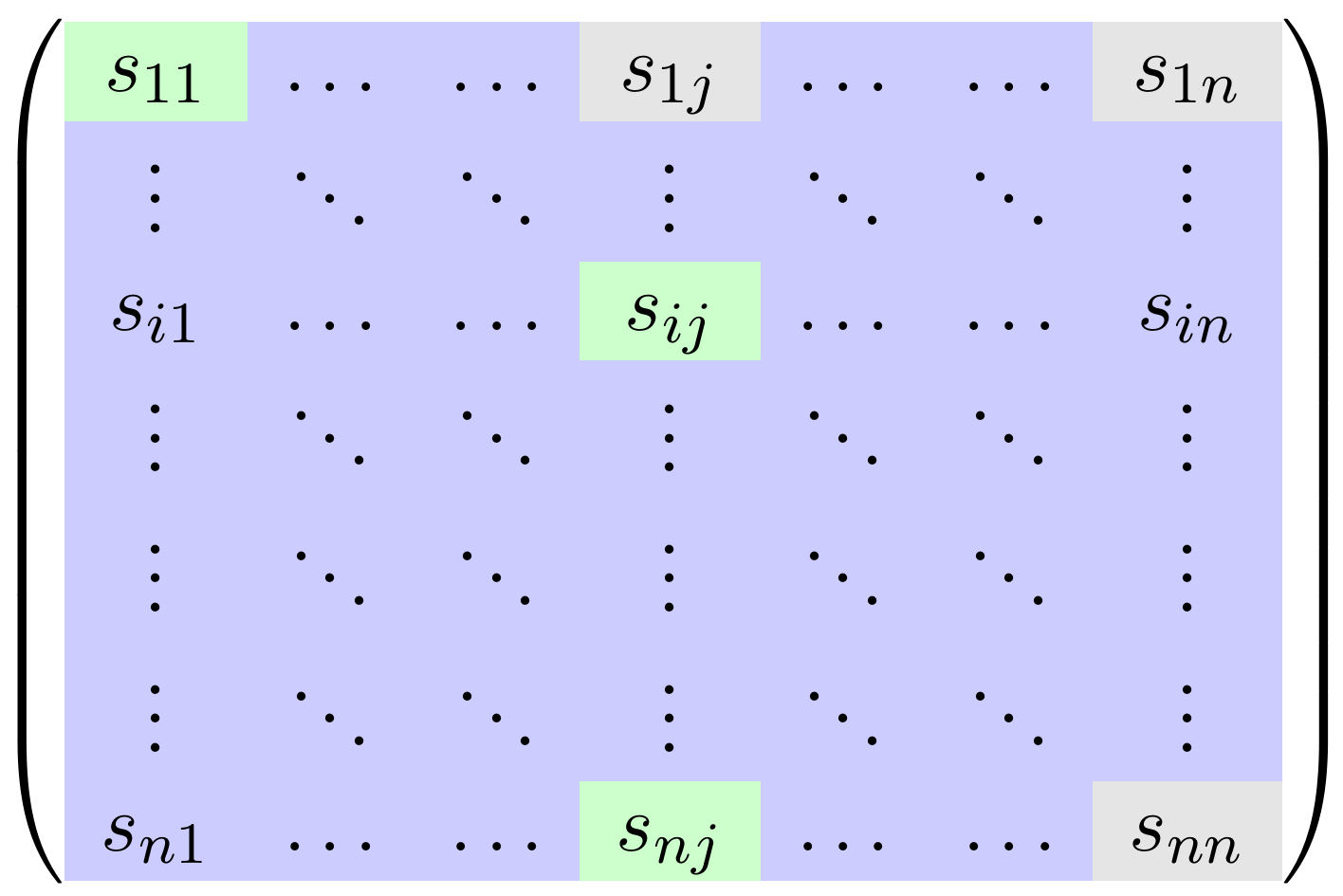}
  \caption{Entity-Pair Ranking protocol (PR)}
  \label{fg:new_protocol}
\end{subfigure}
\caption{Triples considered during the evaluation by both protocols. 
  In ER, the evaluation is performed for each test triple separately. 
  In PR, all test triples for a given relation are considered 
  simultaneously. Notice that the unobserved triple $s_{n1}$ is not a candidate in ER 
  for test triple $s_{ij}$.
  }
\end{figure}

\paragraph{Entity ranking (ER)}
The goal of ER is to assess model performance in terms of
ranking answers to certain questions. In particular, for each test triple $t =
(i, k, j)$, two questions $q_s = (?,k,j)$ and $q_o = (i,k,?)$ are generated. For
question $q_s$, all entities $i' \in \mathcal{E}$ are ranked based on the score
$s(i',k,j)$. To avoid misleading results, entities $i'$ that correspond to
observed triples in the dataset (i.e., $(i',k,j)$ in train/validate) are
discarded to obtain a \emph{filtered ranking}. The same process is applied for
question $q_o$. Model performance is evaluated based on the recorded positions
of the test triples in the filtered ranking. The intuition is that models that
rank test triples (which are known to be true) higher are expected to be
superior. Usually, the micro-average of \textit{filtered Hits@K}---i.e., the
proportion of test triples ranking in the top-$K$---and \textit{filtered
  MRR}---i.e., the mean reciprocal rank of the test triples---are reported.
Figure~\ref{fq:protocol} provides a pictorial view of ER
for a single relation. Given the score matrix of a relation $k$, where $s_{ij}$
is the score of triple $(i, k, j)$, a single test triple is shown 
in green, all candidate triples considered during the evaluation are shown in blue, 
and all triples observed in the training, validation and testing sets (not considered 
during evaluation) are shown in grey.

\subsection{Discussion}\label{sec:discussion}

Regarding triple classification, \citet{DBLP:journals/corr/abs-1709-04808} found that
most models achieve an accuracy of at least 93\%. 
This is due to the fact that negative 
triples with high score are rarely sampled as pseudo-negative triples because of the large
number of entities from which the single replacement entity is picked for a given 
test triple. 
This means that most classification tasks are ``easy''. Consequently, 
the accuracy of triple classification overestimates model performance for KBC tasks. 
This protocol is less adopted in recent work.

We argue that ER also overestimates model performance for KBC. In particular, 
the protocol is more appropriate to evaluate question
answering tasks. Since ER generates questions from true
test triples, it only asks questions that are \emph{known to have an answer}.
The question itself leaks this information from the test data into the
evaluation. This is fine for QA, where the assumption that
only ``sensical'' questions are asked is suitable. In KBC, however, our goal is
to infer missing triples with high precision. Consequently, models need to assess 
every candidate triple, even nonsensical ones such as 
\textit{(Algorithms to Live By, visit, Turing Award)}, because they should be
accurate on these as well.

To better illustrate why ER can lead to misleading results, consider
the DistMult model and the asymmetric relation \textit{nominatedFor}. As
described in Sec.~\ref{sec:preliminaries}, DistMult models all relations as
symmetric in that $s(i,k,j)=s(j,k,i)$. Now consider triple 
$t = \textit{(H. Simon, nominatedFor, Nobel Prize)}$, and let us suppose 
that the model successfully assigns $t$ a high score $s(t)$. Then the inverse 
triple $t'= \textit{(Nobel Prize, nominatedFor, H. Simon)}$ will also obtain 
a high score since $s(t') = s(t)$. Thus, if we use DistMult for KBC, either both 
or none of these triples will be added to the KB: in each case, we make
an error. In ER, however, we never ask a question about $t'$ since there is no 
test triple for this relation containing either \emph{Nobel Prize} as subject or 
\emph{Herbert Simon} as object, so these errors made by DistMult do not affect 
its overall performance. Thus the symmetry property of DistMult does 
not influence the result.

For another example, consider TransE and the lexical relation
$k=\textit{derivationally related form}$, which is symmetric but not
reflexive. One can show that for all $(i,k,j)$, the TransE scores satisfy
\begin{align*}
  &s(i,k,j)+s(j,k,i) \\
  &\quad = -\|\ve_i+\vr_k-\ve_j\|-\|\ve_j+\vr_k-\ve_i\|\\
  &\quad \le -\|\ve_i+\mathbf{0}-\ve_j\|-\|\ve_j+\mathbf{0}-\ve_i\|.
\end{align*}
Since for symmetric relations TransE aims to assign high scores to both
$(i,k,j)$ and $(j,k,i)$, TransE has the tendency to push the relation embedding
$\vr_k$ towards $\mathbf{0}$ as well as $\ve_i$ and $\ve_j$ towards each other.
But when $\vr_k\approx 0$, then $s(i,k,i)$ is large so that the relation is treated as
if it were reflexive. Again, in ER, this property only slighly
influences the results: only one ``'reflexive'' tuple is in each filtered entity
list so that the correct answer $i$ for question $(?,k,j)$ ranks at most one
position lower. The same reasoning applies to question $(i,k,?)$. This can also
be seen in TransE's Hits@1 performance in symmetric relations~\cite{meilicke2018fine}.

In summary, since a high score implies high truthfulness, it is crucial to consider 
the errors introduced by negative triples with 
high scores when evaluating KBC performance, otherwise we could get a misleading 
impression of a model's precision. The current evaluation protocols may systematically 
ignore high-scored false triples, thereby potentially overestimating model performance.

\subsection{Entity-Pair Ranking Protocol}\label{sec:epr}

We propose an alternative protocol 
called Entity-Pair Ranking (PR). While PR may not be the final solution for automatic evaluation,  
it is still helpful to answer our research question, as we will see in the next section.
PR is simple: for each relation $k$, we ask question $(?,k,?)$. As before, 
we use the model to rank all answers,
i.e., pairs of entities, and filter out training and validation data in the
ranking so as to rank only triples not used during model training. In this way,
any negative triples with a high score will appear at a top position, making it
harder for true triples to rank high. 
Figure~\ref{fg:new_protocol} shows the contrast between the number of negative
triples considered for entity-pair and those considered for ER. Again,
test triples are shown in green, candidate triples are shown in blue,
and triples observed during training and validation are shown in grey. 
The number of candidates is much higher than those considered for ER. However,
when answering the question $(?,k,?)$ with all possible entity pairs, all test triples 
for relation $k$ will be ranked simultaneously. 
Let $|\mathcal{T}_{k}|$ be the number of test triples in $k$. 
ER needs to consider in total $2\cdot|\mathcal{T}_k|\cdot|\cE| - 1$ candidates for $k$, 
while PR needs to consider $|\cE|^2$ candidates.

Since all test triples in relation $k$ are considered at once, we do not rely on 
MRR for PR, but consider weighted \textit{MAP@K}, i.e., weighted mean average 
precision in the top-$K$ filtered results, and weighted \textit{Hits@K}, i.e., 
weighted percentage of test triples in the top-$K$ filtered results. For a fixed 
$K$, the weights for each relation $k$ are proportional to the number of test triples 
$|\cT_k|$ and upper bounded by $K$: 
\begin{align}
\text{MAP}@K &= \sum\limits_{k\in \mathcal{R}} \text{AP}_k@K\times \frac{ \min(K, |\mathcal{T}_k|)}{\sum\limits_{k' \in
  \mathcal{R}} \min(K, |\mathcal{T}_{k'}|)} \nonumber \\
\text{Hits}@K &= \sum\limits_{k\in \mathcal{R}} \text{Hits}_k@K\times \frac{ \min(K, |\mathcal{T}_k|)}{\sum\limits_{k' \in
  \mathcal{R}} \min(K, |\mathcal{T}_{k'}|)} \nonumber
\end{align}
where $\text{AP}_k@K$ is the average precision of the top-$K$ list with respect 
to the test tuples and $\text{Hits}_k@K$ corresponds to the ratio of test triples 
in the top $K$ (i.e., $\text{(\# test triples in top-$K$})/\min(K,|\mathcal{T}_k|)$).
In a nutshell, infrequent relations are not treated the same as frequent ones during
evaluation due to the weighting factor.

Note that all evaluation methods for KBC may underestimate model performance. 
This happens when models rank unobserved true triples 
(neither in train/validate/test) high. This behaviour is generally unavoidable for any 
protocol without further background knowledge or manual labelling, but it may be of
particular importance in this protocol due to the large number of candidates considered.
We empirically study this underestimation effect in 
PR in Sec.~\ref{sec:decode}.

Another concern about the entity-pair ranking
protocol is its computation cost. There are $\left| \cE \right|^2$
possible entity pairs to consider per relation, and it may be infeasible to
compute the score of all these pairs when there is a large number of entities.
Note, however, that the protocol only makes use of the high-scoring entity
pairs. Such entity pairs may be determined more efficiently, e.g., using
techniques for maximum inner-product search such as
\citet{DBLP:conf/nips/Shrivastava014} or by exploiting available background
knowledge as in Sec.~\ref{sec:bk}. 
Moreover, even when the scores of all entity pairs have to be computed, 
one does not have to sort all the scores to obtain the sorted top-$K$.
For example, one could use Quickselect to obtain the $K$ highest scores and then sort
only these.
Although we did not run into performance
issues in our experimental study, more work into efficient retrieval methods is
needed to support larger datasets. 
Notice that the confidence of an embedding model of a candidate triple 
is expressed by the score of that triple. If an embedding model cannot determine which 
triples are high-scoring, it is unclear how to produce candidate triples to add to the KB. 
Thus, in general, if we cannot determine
high-scoring triples from a particular embedding model, then that embedding
model may not be suitable for KBC in the first place.
Following this argument, a metric based on only top-$K$ candidates suits better than a 
holistic one based on the ranking of all candidates for KBC, since a model which predicts 
well for top-$K$ but performs weakly in terms of the whole ranking is still useful for KBC.
\section{Experimental Study}\label{sec:experiments}

\begin{table}
 \centering
  \begin{tabular}{lrrrrr}
    \hline
    Dataset   & $|\mathcal{E}|$ & $|\mathcal{R}|$ & $|\mathcal{T}^{train}|$ & $|\mathcal{T}^{val}|$ & $|\mathcal{T}^{test}|$    \\ \hline
    FB15K     & \n{14951}       & \n{1345}        & \n{483142}              & \n{50000}               & \n{59071} \\
    FB-237    & \n{14505}       & \n{237}         & \n{272115}              & \n{17535}               & \n{20466} \\ 
    WN18      & \n{40943}       & \n{18}          & \n{141442}              & \n{5000}                & \n{5000}  \\
    WNRR      & \n{40559}       & \n{11}          & \n{86835}               & \n{2824}                & \n{2924}  \\ \hline
  \end{tabular}
  \caption{Dataset statistics}
  \label{tab:data}
\end{table}

We conducted experiments to study the performance of various
embedding models for KBC. As baseline, we consider a very simple 
system called \textit{RuleN}~\cite{meilicke2018fine}, as it has shown 
good performance according to the existing entity ranking results.
Our goal was not to determine which model currently works best, 
but rather to determine whether they are suitable for KBC when asking the
question $(?, k, ?)$.
In addition, we investigated the extent to which the underestimation effect 
due to unobserved triples affects PR. Finally, to determine whether the results 
provided by PR were mostly due to the high number of nonsensical triples, 
we used background knowledge to study the performance of models when 
filtering out most nonsensical triples. All datasets, experimental results, 
and source code will be made publicly available.
 

\begin{table*}
\centering
\begin{tabular}{p{1.1cm}S[table-column-width=1.6cm,table-format=2.2]S[table-column-width=1.6cm,table-format=2.2]S[table-column-width=1.6cm,table-format=2.2]S[table-column-width=1.6cm,table-format=2.2]S[table-column-width=1.6cm,table-format=2.2]S[table-column-width=1.6cm,table-format=2.2]S[table-column-width=1.6cm,table-format=2.2]S[table-column-width=1.6cm,table-format=2.2]S[table-column-width=1.6cm,table-format=2.2]}
     \hline
     Dataset  & \multicolumn{2}{c}{FB15K}   & \multicolumn{2}{c}{FB-237}  & \multicolumn{2}{c}{WN18}    & \multicolumn{2}{c}{WNRR}    \\ \hline
     Model    & {MRR (\%)}  & {Hits@K (\%)} & {MRR (\%)}  & {Hits@K (\%)} & {MRR (\%)}  & {Hits@K (\%)} & {MRR (\%)}  & {Hits@K (\%)} \\ \hline
     DistMult & 66.0        & 84.5            & 27.0        & 43.2        & 79.0        & 93.7          & 43.2        & 47.4          \\
     TransE   & 50.0        & 77.7            & 29.0        & 46.6        & 72.0        & 90.8          & 22.0        & 49.1          \\
     ComplEx  & 67.2        & 83.3            & 28.0        & 43.5        & 94.0        & 94.8          & 44.0        & 48.1          \\
     Analogy  & 67.0        & 83.8            & 27.0        & 43.3        & 94.1        & 94.2          & 44.0        & 48.6          \\
     RESCAL   & 44.4        & 68.7            & 27.0        & 42.7        & 92.0        & 93.9          & 42.0        & 44.7          \\ \hline 
     RuleN    & 80.5        & 87.0            & 26.0        & 42.0        & 95.0        & 95.8          & {---}       & 53.6          \\ \hline
\end{tabular}
\caption{Results with the commonly used entity ranking protocol (ER) ($K=10$)}
\label{tab:entity_ranking}
\end{table*}

\begin{table*}
\centering
\begin{tabular}{p{1.1cm}S[table-column-width=1.6cm,table-format=2.2]S[table-column-width=1.6cm,table-format=2.2]S[table-column-width=1.6cm,table-format=2.2]S[table-column-width=1.6cm,table-format=2.2]S[table-column-width=1.6cm,table-format=2.2]S[table-column-width=1.6cm,table-format=2.2]S[table-column-width=1.6cm,table-format=2.2]S[table-column-width=1.6cm,table-format=2.2]S[table-column-width=1.6cm,table-format=2.2]}
    \hline
    Dataset   & \multicolumn{2}{c}{FB15K} & \multicolumn{2}{c}{FB-237} & \multicolumn{2}{c}{WN18}   & \multicolumn{2}{c}{WNRR}                \\ \hline
    Model     & {MAP@K (\%)}  & {Hits@K (\%)} & {MAP@K (\%)}  & {Hits@K (\%)} & {MAP@K (\%)}  & {Hits@K (\%)} & {MAP@K (\%)}  & {Hits@K (\%)} \\  \hline
    DistMult  & 1.3           & 10.4          & 0.3           & 4.2           & 7.9           & 9.7           & 14.1          & 17.8          \\
    TransE    & 21.1          & 36.3          & 7.9           & 17.6          & 22.3          & 31.5          & 0.2           & 1.3           \\
    ComplEx   & 25.9          & 45.2          & 7.1           & 16.6          & 78.5          & 87.7          & 16.8          & 20.0          \\
    Analogy   & 18.8          & 34.8          & 4.9           & 14.3          & 61.5          & 75.8          & 15.4          & 19.8          \\
    RESCAL    & 15.0          & 30.3          & 6.7           & 15.0          & 48.2          & 60.9          & 13.1          & 13.8          \\ \hline
    RuleN     & 77.4          & 83.7          & 7.6           & 15.8          & 94.8          & 96.8          & 21.5          & 25.1          \\ \hline
\end{tabular}
\caption{Results with the proposed entity-pair ranking protocol (PR) ($K=100$)}
\label{tab:new_evaluation-results}
\end{table*}

\subsection{Experimental Setup}

\paragraph{Datasets.} We use four common KBC benchmark datasets: FB15K, WN18, 
FB-237, and WNRR. The first two are subsets of WordNet and Freebase,
respectively~\cite{DBLP:conf/nips/BordesUGWY13}, and it is known that the 
regularities in the datasets can be explained via simple implication 
rules~\cite{dettmers2018conve,meilicke2018fine}. We will see that despite 
their simplicity, these datasets can still shed some light on the behavior 
of various models under PR. FB-237 is constructed from FB15K to make the dataset 
more challenging~\cite{DBLP:conf/emnlp/ToutanovaCPPCG15}, and model performance is
indeed considerably lower~\cite{akrami2018re}. Let $\mathcal{T}^{train},
\mathcal{T}^{val}, \mathcal{T}^{test}$ be the training, validation, and test
data, resp. Then FB-237 is obtained from FB15k by removing inverse relations and
by ensuring that whenever $(i,k,j)\in \mathcal{T}^{test} \cup \mathcal{T}^{val}$,
$(i,k',j)\notin \mathcal{T}^{train}$ for all $k'\neq k$. WNRR is constructed from 
WN18 by following similar procedure. Key dataset statistics are summarized in 
Table~\ref{tab:data}. Interestingly, the latter two datasets were created when
considering the former two datasets unsuitable for the task. Conversely, we suggest
that independent of the dataset, the standard evaluation protocol in unsuitable.

\paragraph{Negative sampling.}
Since we are only given true but no false triples, embedding models are usually
trained using a negative sampling strategy to obtain pseudo-negative 
triples~\cite{nickel2016review}. We consider three sampling strategies in our 
experiments: \\
\textit{Perturb 1}: For each training triple $t=(i,k,j)$, sample each
pseudo-negative triple by randomly replacing either $i$ or $j$ by a random
entity (such that the resulting triple is unobserved). This ensures that at
least one entity is observed in the training data of relation $k$, which to some
extent avoids nonsensical pseudo-negative triples. This sampling is commonly used 
and matches ER, which is based on questions $(?,k,j)$ 
and $(i,k,?)$. \\
\textit{Perturb 2}: For each training triple $t=(i,k,j)$, sample pseudo-negative
triples by sampling random unobserved tuples from relation $k$. This
method produces many nonsensical negative triples, but is more suited to
KBC, which is based on the question $(?,k,?)$. \\
\textit{Perturb 1-R}: For each training triple $t=(i,k,j)$, sample each
pseudo-negative triple by randomly replacing either $i$, $k$ or $j$ by a random
entity (or relation for $k$). The generated negative samples are not compared with
the training set~\cite{DBLP:conf/icml/LiuWY17}.

\paragraph{Training and implementation.} We trained all embedding models as in previous 
work~\cite{DBLP:conf/icml/TrouillonWRGB16,DBLP:conf/icml/LiuWY17}.
In particular, we used AdaGrad~\cite{DBLP:journals/jmlr/DuchiHS11} 
as an optimizer and trained DistMult, ComplEx, Analogy and RESCAL 
with binary cross-entropy loss, which works well in practice. 
TransE always produces negative scores, so it is unclear how to 
train it with cross-entropy loss in a principled way. We thus used 
pair-wise ranking loss with margin $\gamma\ge 0$ as 
in~\citet{DBLP:conf/nips/BordesUGWY13}. We implemented all embeddding models on top of 
the code from~\citet{DBLP:conf/icml/LiuWY17}\footnote{https://github.com/quark0/ANALOGY}
in C++ using OpenMP. 
For RuleN, we use the original implementation provided by the authors.
The evaluation protocols were written in Python, with
Bottleneck{\footnote{https://pypi.org/project/Bottleneck/} used for efficiently obtaining
the top K entries in a given score matrix, such that only these entries required sorting.

\paragraph{Hyperparameters.} 
The best hyperparameters are selected based on MRR (for ER) and
MAP@100 (for PR) on the validation data. These are reported for both evaluation
protocols in the appendix. For both protocols, we performed an exhaustive grid 
search over the following hyperparameter settings:
$d \in \{100,150,200\}$,
weight of $l_2$-regularization $\lambda \in \{0.1, 0.01, 0.001\}$, 
learning rate $\eta \in \{0.01,0.1\}$,
negative sampling strategies \emph{Perturb 1}, \emph{Perturb 2} and \emph{Perturb 1-R}, 
and margin hyperparameter $\gamma \in \{0.5,1,2,3,4\}$ for TransE. For each training 
triple, we sampled $6$ pseudo-negative triples. To keep effort tractable, we only used
the most frequent relations from each dataset for hyperparameter tuning (top-5, top-5,
top-15, and top-30 most frequent relations for WN18, WNRR, FB-237 and FB-15k
respectively).\footnote{All datasets are highly skewed.} We trained 
each model for up to 500 epochs during grid search (1800 for TransE).
\footnote{Chosen such that models are usually converged.}.
In all cases, we used early stopping, i.e., we evaluated model performance every 50 epochs 
and used the overall best-performing model.
The best hyperparameters for both evaluation protocols are reported in the 
online material. We found that \textit{Perturb-2} can indeed be useful in both 
protocols. 

RuleN learns path rules of a given length and a particular type of constant rules.
Moreover, RuleN takes a sample of the KB when looking for groundings of rules.
We refer to ~\citet{meilicke2018fine} for the details. We use the best settings 
reported by the authors with respect to entity ranking~\cite{meilicke2018fine}. 
That is, for FB15K and FB-237, we learned path rules of length 2 and constant rules, 
and the sampling size was set to 1000. For WN18 and WNRR, we learned path rules of 
length 2 with sampling size of 1000, and path rules of length 3 with sampling size of 100, 
as well as constant rules. For PR, we learned path rules of length 2 using a sampling size of 
500 for FB15K and FB-237. For WN18 and WNRR, we learned path rules of length 3 and set 
the sampling size to 500.

\subsection{Performance Results with ER}

Table~\ref{tab:entity_ranking} summarizes our results with ER. 
Embedding models perform competitively with respect to RuleN on all datasets,
except for their MRR performance on FB15K. Notice that this generally holds even for the 
more restricted models (TransE and DistMult) on the more challenging datasets, which were 
created after criticizing FB15K and WN18 as too 
easy~\cite{toutanova2015observed,dettmers2018conve}.
Due to such results, DistMult is considered a state-of-the-art model for 
KBC~\cite{DBLP:conf/rep4nlp/KadlecBK17}.
This seems counterintuitive (especially considering the more difficult 
datasets), since DistMult can only model symmetric relations, but most 
relations in these datasets are asymmetric. Regarding TransE,
while its MRR was relatively low on all datasets except FB-237, it still achieved great
performance in Hits@10. Again, this is counterintuitive because WN18 
contains a large number of symmetric relations, so it should be difficult for 
TransE to achieve good results. These observations indicate that ER may not be 
well-suited for KBC.

\subsection{Performance Results with PR}\label{sec:exp_new_protocol}

The evaluation results of PR with $K = 100$ are summarized in
Table~\ref{tab:new_evaluation-results}.  Note that Tables~\ref{tab:entity_ranking} 
and \ref{tab:new_evaluation-results} are not directly comparable, because PR considers
all test triples for a given relation simultaneously. Consequently, it is more difficult 
to rank at the top in PR, so we chose $K > 10$. Performance for different values of $K$
is reported in Fig.~\ref{fg:hits_curves}. Also, recall that for ER we compute
MRR, while for PR we compute MAP, as explained in Sec.~\ref{sec:epr}. The effect of 
the choice of $K$ is discussed later in this section. 

For now, assume that there is no unobserved true triples other than the test set. 
We defer the discussion of this issue in Sec.~\ref{sec:decode}. Since knowledge 
bases strive for high accuracy, we consider models unsatisfactory if we cannot 
use them to add new facts without sacrificing this accuracy throughout this section.
For the embeddings, observe that the performance of all models is unsatisfactory on all 
datasets, especially when compared with RuleN on FB15K and WN18, which were previously
considered to be too easy for embedding models.
Specifically, DistMult's Hits@100 is slightly less than 10\% on WN18, meaning that if we add
the top 100 ranked triples to the KB, over 90\% of what is added is likely false. 
Even when using ComplEx, the best model on FB15K, we would potentially add more than 
50\% false triples. 
This implies that embedding models cannot  even capture simple rules successfully.
The notable exceptions are Analogy and ComplEx on WN18. TransE and DistMult did not 
achieve competitive results on WN18, even when compared with other embedding models. 
In addition, DistMult did not achieve competitive results on FB15K and FB-237 and 
TransE did not achieve competitive results in WNRR. When considering all datasets, 
there was no single embedding model which outperformed all others. In general, ComplEx 
and Analogy achieved consistently better results across different datasets than
other models. But when compared with the baseline, even the performance of these models was 
often not satisfactory. This suggests that better models and/or training strategies are
needed for embedding models.

\begin{figure}
  \centering
  \begin{tabular}{m{6.5cm}m{1.2cm}}
    \includegraphics[scale=1.22]{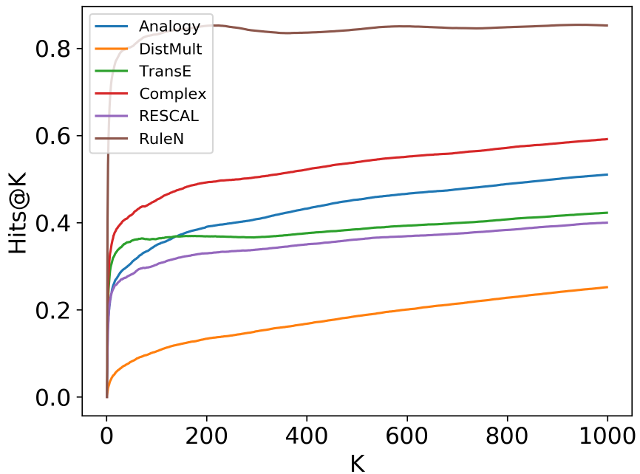}      & FB15K   \\
    \includegraphics[scale=1.22]{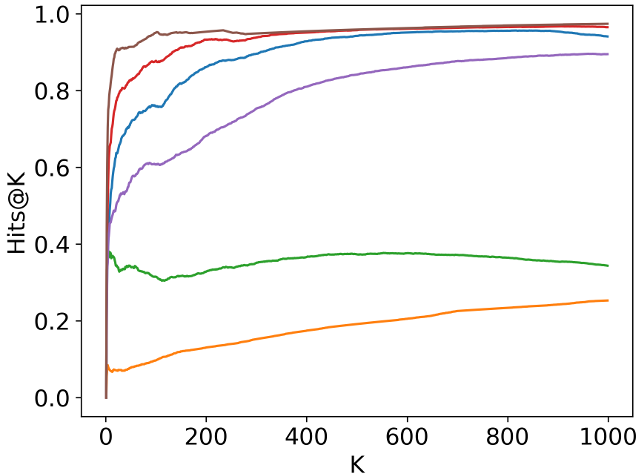}    & WN18    \\
    \includegraphics[scale=1.22]{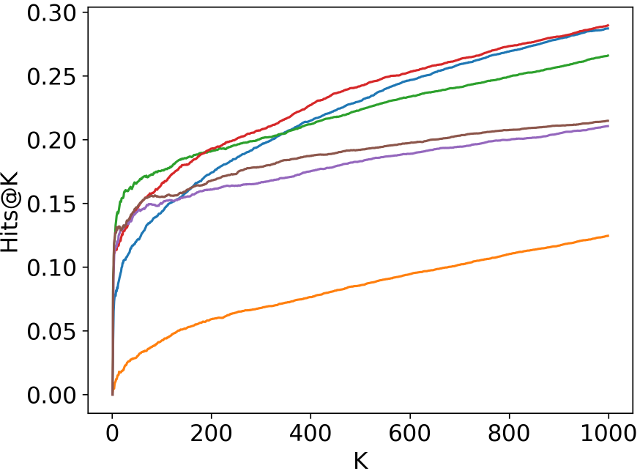}  & FB-237  \\
    \includegraphics[scale=1.22]{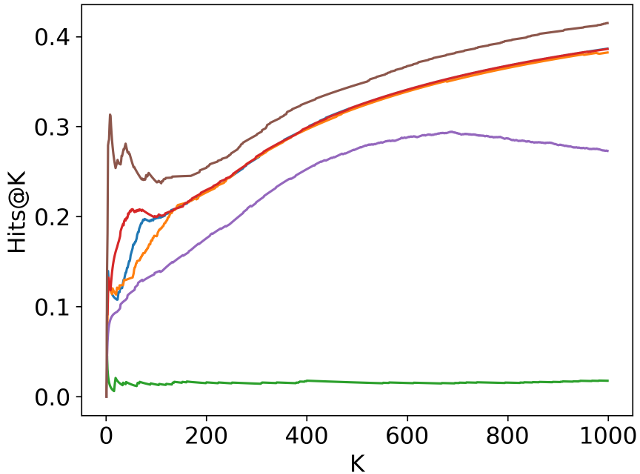}     & WNRR
  \end{tabular}
  \caption{Hits@K with PR as a function of K}
  \label{fg:hits_curves}
\end{figure}

\begin{figure}
  \centering
  \begin{tabular}{m{6.5cm}m{1.2cm}}
    \includegraphics[scale=1.22]{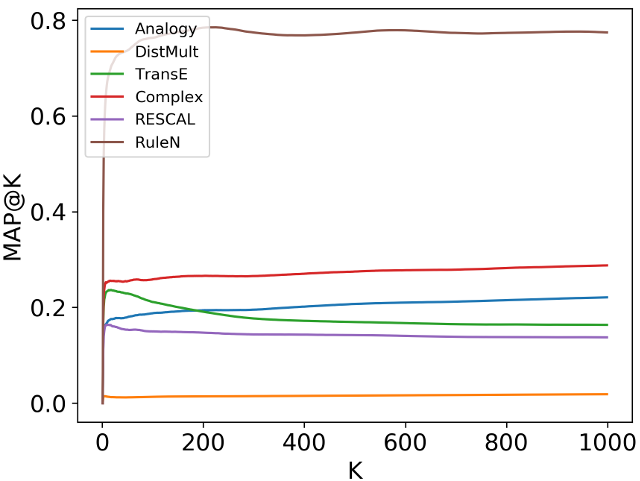}      & FB15K   \\
    \includegraphics[scale=1.24]{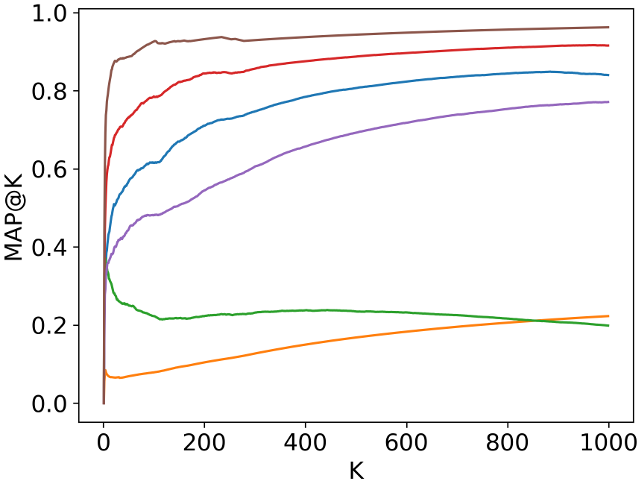}    & WN18    \\
    \includegraphics[scale=1.24]{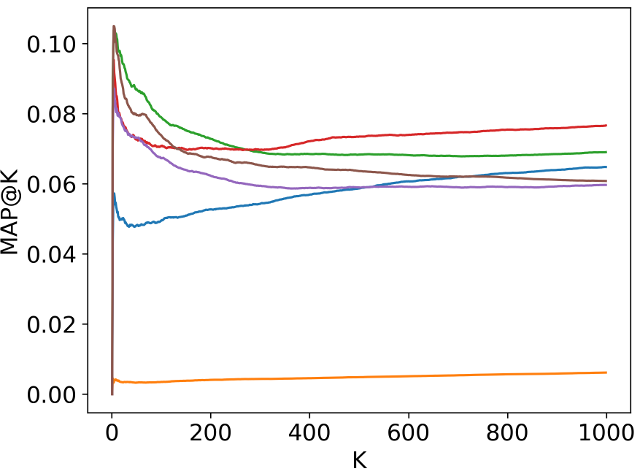}  & FB-237  \\
    \includegraphics[scale=1.24]{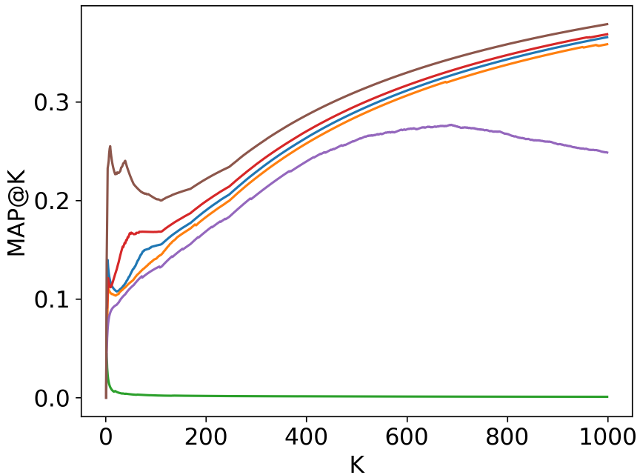}     & WNRR
  \end{tabular}
  \caption{MAP@K with PR as a function of K}
  \label{fg:map_curves}
\end{figure}

With respect to FB-237 and WNRR, it is intrinsically difficult for rule-based 
approaches to learn simple rules with high confidence because of the way these datasets were 
constructed~\cite{meilicke2018fine}. As a result, RuleN does not 
perform well on these two datasets. This is reflected both by ER and PR. 


\begin{table*}
\centering
\begin{tabular}{l|S[table-column-width=0.8cm,table-format=4.0]S[table-column-width=0.8cm,table-format=4.0]S[table-column-width=0.8cm,table-format=4.0]S[table-column-width=0.8cm,table-format=4.0]S[table-column-width=0.8cm,table-format=4.0]S[table-column-width=0.8cm,table-format=4.0]S[table-column-width=0.8cm,table-format=4.0]S[table-column-width=0.8cm,table-format=4.0]S[table-column-width=0.8cm,table-format=4.0]S[table-column-width=0.8cm,table-format=4.0]S[table-column-width=0.8cm,table-format=4.0]S[table-column-width=0.8cm,table-format=4.0]}
\hline
\multicolumn{1}{l|}{}                   & \multicolumn{12}{c}{Model}                                                                                                                                                            \\ \cline{2-13}
Relation                                & \multicolumn{2}{c}{DistMult}      & \multicolumn{2}{c}{TransE}  &  \multicolumn{2}{c}{ComplEx} & \multicolumn{2}{c}{Analogy} & \multicolumn{2}{c}{RESCAL} & \multicolumn{2}{c}{RuleN} \\ \hline
\textit{hyponymy}                       & 1   & {(}1{)}                     & 18  & {(}32{)}              & 99  & {(}99{)}              & 99  & {(}99{)}              & 92  & {(}93{)}              & 100  & {(}100{)}          \\
\textit{hypernymy}                      & 0   & {(}0{)}                     & 5   & {(}33{)}              & 99  & {(}99{)}              & 99  & {(}99{)}              & 96  & {(}98{)}              & 100  & {(}100{)}          \\
\textit{derivationally related form}    & 100 & {(}100{)}                   & 0   & {(}0{)}               & 100 & {(}100{)}             & 100 & {(}100{)}             & 6   & {(}68{)}              & 100  & {(}100{)}          \\
\textit{member meronym}                 & 0   & {(}0{)}                     & 18  & {(}41{)}              & 74  & {(}84{)}              & 83  & {(}85{)}              & 44  & {(}63{)}              & 100  & {(}100{)}          \\
\textit{member holonym}                 & 0   & {(}0{)}                     & 16  & {(}47{)}              & 74  & {(}83{)}              & 83  & {(}85{)}              & 37  & {(}54{)}              & 100  & {(}100{)}          \\  \hline
\end{tabular}
\caption{Number of test triples in the top-100 filtered predictions on WN18.
    An estimate of the number of true triples in the top-100
  list is given in parentheses.}
\label{tab:inspect}
\end{table*}

To better understand the behavior of TransE and DistMult, we investigated their
performance on the top-5 most frequent relations on WN18.
Table~\ref{tab:inspect} shows the number of test triples appearing in the
top-100 for each relation (after filtering triples from the training and
validation sets). The numbers in parentheses are discussed in
Section~\ref{sec:decode}.

We found that DistMult worked well on the symmetric relation
\textit{derivationally related form}, where its symmetry assumption clearly
helps. Here 93\% of the training data consists of symmetric pairs (i.e.,
$(i,k,j)$ and $(j,k,i)$), and 88\% of the test triples have its symmetric
counterpart in the training set. In contrast, TransE finds no test triples for
\textit{derivationally related form} in the top-100 list. We found that the
norm of the embedding vector of this relation was $0.1$, which was considerably
smaller than for the other relations (avg.~$1.4$). This supports our argument 
that TransE has a tendency to push symmetric relations embeddings to $\mathbf{0}$.

Note that while \textit{hyponymy}, \textit{hypernymy}, \textit{member meronym}
and \textit{member holonym} are semantically transitive, the dataset contains
almost exclusively their transitive core, i.e., the dataset (both train and test) 
does not contain many of the transitive links of the relations. As a result, the
models do not ``see'' their transitivity. Thus, models that cannot handle 
transitivity well may still produce good results. This might explain why 
TransE performs better for these relations than for \textit{derivationally related form}. 
DistMult did not perform well on these relations (they are asymmetric). ComplEx and Analogy 
showed superior performance across all relations. RESCAL 
is in between, most likely due to difficulties in finding a good parameterization. 
However, it is unclear why TransE performs well on FB15K and FB-237.

To investigate the performance of models for different values of $K$, we give
the curves of Hits@$K$ and MAP@$K$ as a function of $K$ for all datasets in 
Fig.~\ref{fg:hits_curves} and Fig.~\ref{fg:map_curves} respectively.
ComplEx and Analogy performed best for large $K$ w.r.t. other embedding models. 
Similarly, TransE works the best for small values of $K$ on FB15K and FB-237.
Notice that RuleN performs considerably better on FB15K, WN18 and WNRR, while
it still performs competitively on FB-237.

Regarding runtimes, we found that PR is 3 to 4 times slower than ER, 
but still manageable for these datasets (30$\sim$90 minutes in general).


\subsection{Influence of Unobserved True Triples}\label{sec:decode}

Since all datasets are based on incomplete knowledge bases, evaluation protocols may systematically 
underestimate model performance. For ranking-based evaluations in particular, a true triple that is 
neither in the training, nor validation, nor test data may be ranked high by some model, but we treat 
that triple as negative during the evaluation. Such models would be penalized. In fact, all prior 
evaluation protocols also have this issue, but PR might be particularly sensitive to this due to the 
large number of candidates considered during evaluation.

Generally, it is unclear how to design an automatic evaluation strategy that avoids this problem. 
Manual labeling can be used to address this, but it may sometimes be infeasible given the large number 
of relations, entities, and models for KBC. Moreover, depending on the domain of the data, such 
labeling may even require expert knowledge.

To explore such underestimation effect in PR, we decoded the unobserved triples in the top-100 
predictions of the 5 most frequent relations of WN18. We then checked whether those triples are 
implied by the symmetry and transitivity properties of each relation.\footnote{To do so, we 
computed the symmetric/transitive closure of the available data in the entire WordNet knowledge 
base.} In~Table~\ref{tab:inspect}, we give the resulting number of triples in
parentheses (i.e., number of test triples + implied triples). We observed that
underestimation indeed happened. TransE was mostly affected, but still did not
lead to competitive results when compared to ComplEx and Analogy. RuleN achieves the best possible
results in all 5 relations. These results suggest that (1) underestimation is indeed a concern, 
and (2) the results reported by the PR can nevertheless give an indication of relative model performance.

\subsection{Type Filtering}\label{sec:bk}

When background knowledge (BK) is available, embedding models only need to predict
triples consistent with this BK. Notice that this is inherently what rule-based approaches
do, since all predicted candidates will be type-consistent.
We explored whether their 
performance can be improved by filtering out nonsensical triples from each model's predictions.
This should also help reduce computational costs. In 
particular, we investigated how model performance is affected when
we filter out predictions that violate type constraints (domain and range of
each relation). If a model performance improves with such type filtering, it
must have ranked tuples with incorrect types high in the first place. We can thus
assess to what extent models capture entity types as well as the domain and
range of the relations.

We extracted from Freebase
type definitions for entities and domain and range constraints for relations. 
We also added the domain (or range) of a relation $k$ to the type set of each subject 
(or object) entity which appeared in $k$.
We obtained types for all entities in both FB datasets, and domain/range specifications 
for roughly 93\% of relations in FB15K and 97\% of relations in FB-237. The remaining 
relations were evaluated as before.

\begin{table}
  \centering
    \begin{tabular}{llrrrr}
    \hline
      Data  & Model     & \multicolumn{2}{c}{MAP@K (\%)}  & \multicolumn{2}{c}{Hits@K (\%)} \\ \hline
      FB15K & DistMult  & 18.8         & (+17.5)          & 36.4        & (+26.0)           \\
            & TransE    & 25.7         & (+4.5)           & 41.7        & (+5.4)            \\
            & ComplEx   & 41.8         & (+15.9)          & 61.9        & (+16.7)           \\
            & Analogy   & 41.3         & (+22.5)          & 61.5        & (+26.7)           \\
            & RESCAL    & 16.7         & (+1.7)           & 32.8        & (+2.5)            \\ \hline
            & RuleN     & 77.4         & (0.0)            & 83.7        & (0.0)             \\ \hline
     FB-237 & DistMult  & 9.5          & (+9.2)           & 18.1        & (+13.9)           \\
            & TransE    & 11.3         & (+3.4)           & 21.2        & (+3.6)            \\
            & ComplEx   & 11.3         & (+4.2)           & 21.8        & (+5.2)            \\
            & Analogy   & 10.5         & (+5.6)           & 20.9        & (+6.6)            \\
            & RESCAL    & 10.2         & (+3.5)           & 19.0        & (+4.0)            \\  \hline
            & RuleN     & 7.6          & (0.0)            & 15.8        & (0.0)             \\  \hline
  \end{tabular}
  \caption{Results with PR using type filtering (K = 100).}
  \label{tab:schema-results}
\end{table}

We report in Table~\ref{tab:schema-results} the Hits@100 and MAP@100
as well as their absolute improvement (in parentheses) w.r.t.
~Table~\ref{tab:new_evaluation-results}. We include the results of RuleN 
from Table~\ref{tab:new_evaluation-results}, since
rule-based systems will never predict type-inconsistent facts.
The results show that all models improve by type filtering; thus all models 
do predict triples with incorrect types. In particular, DistMult shows 
considerable improvement on both datasets. Indeed, about 90\% of the relations 
in FB15K (about 85\% for FB-237) have a different type for their domain and 
range. As DistMult treats all relations as symmetric, it introduces a wrong 
triple for each true triple into the top-$K$ list on these relations; type 
filtering allows us to ignore these wrong tuples. This is also consistent with 
DistMult's improved performance under ER, where type constraints are implicitly 
used since only questions with correct types are considered. We also observed 
that ComplEx and Analogy improved considerably on FB15K. Although it is unclear 
why this is the case, it implies that the best performing models on this dataset
according to ER, are still making a considerable number of type-inconsistent
predictions. On FB15K, the relative ranking of the models with type filtering 
is roughly equal to the one without type filtering. On the harder FB-237 dataset, 
all models now perform similarly. Notice that when compared with the performance
of RuleN, embedding models are still far behind on FB15K, but are no longer behind
on FB-237.

Finally, due to the reduction in the number of candidates, the average evaluation 
time was reduced to about 30\% of the time required without BK.

\section{Conclusion}\label{sec:conclusion}

We investigated the question of whether current embedding models provide good
results for the knowledge base completion task. We found that the entity ranking 
evaluation protocol that is currently widely used is tailored to question answering, but
may give a misleading picture of each model's performance with respect to
knowledge base completion, where the aim is high precision. We tested a new and simple 
evaluation protocol with KBC in mind and evaluated a number of state-of-the-art 
models under this protocol. We found that most models did not produce satisfactory results,
especially when compared against a simple rule-based system used as baseline.
This suggests that more research into embedding models and training methods is needed to 
assess whether, when, and how KB embedding models can provide high-quality results.

\clearpage
\bibliographystyle{ACM-Reference-Format}
\bibliography{references}

\clearpage

\section*{Appendix}

\subsection{Hyperparameters of best models}

\begin{table}[!h]
  \centering
    \begin{tabular}{llrS[table-format=1.2]S[table-format=1.3]S[table-format=1.1]c}
    \hline
   Data & Model           & $d$   & $\eta$  & $\lambda$   & $\gamma$ & Perturb    \\ \hline
  FB15K & DistMult        & 200   & 0.1     & 0.01        & {-}      & 1-R        \\
        & TransE\_$l_1$   & 100   & 0.1     & {-}         & 2.0      & 1          \\ 
        & ComplEx         & 200   & 0.1     & 0.001       & {-}      & 1-R        \\
        & Analogy         & 200   & 0.1     & 0.001       & {-}      & 1-R        \\  
        & RESCAL          & 200   & 0.1     & 0.1         & {-}      & 1          \\ \hline
 FB-237 & DistMult        & 200   & 0.1     & 0.01        & {-}      & 2          \\
        & TransE\_$l_1$   & 100   & 0.1     & {-}         & 4.0      & 1          \\ 
        & ComplEx         & 200   & 0.1     & 0.01        & {-}      & 2          \\
        & Analogy         & 200   & 0.1     & 0.01        & {-}      & 2          \\  
        & RESCAL          & 150   & 0.1     & 0.001       & {-}      & 2          \\  \hline
   WN18 & DistMult        & 100   & 0.1     & 0.001       & {-}      & 1          \\
        & TransE\_$l_1$   & 150   & 0.1     & {-}         & 4.0      & 1-R        \\ 
        & ComplEx         & 150   & 0.1     & 0.1         & {-}      & 1          \\
        & Analogy         & 200   & 0.1     & 0.001       & {-}      & 1-R        \\  
        & RESCAL          & 100   & 0.1     & 0.01        & {-}      & 1          \\ \hline
   WNRR & DistMult        & 200   & 0.01    & 0.001       & {-}      & 1          \\
        & TransE$_{l_1}$  & 200   & 0.01    & {-}         & 2.0      & 2          \\
        & ComplEx         & 200   & 0.01    & 0.01        & {-}      & 1          \\
        & Analogy         & 200   & 0.01    & 0.01        & {-}      & 1          \\
        & RESCAL          & 150   & 0.1     & 0.1         & {-}      & 1          \\ \hline
  \end{tabular}
  \caption{Hyperparameters for best models in ER}
  \label{tab:hyper-classic}
\end{table}

\begin{table}[!h]
  \centering
    \begin{tabular}{llrS[table-format=1.2]S[table-format=1.3]S[table-format=1.1]c}
    \hline
   Data & Model           & $d$   & $\eta$  & $\lambda$   & $\gamma$ & Perturb    \\ \hline
  FB15K & DistMult        & 200   & 0.1     & 0.001       & {-}      & 2          \\
        & TransE$_{l_1}$  & 150   & 0.1     & {-}         & 3.0      & 1          \\
        & ComplEx         & 200   & 0.1     & 0.01        & {-}      & 1-R        \\
        & Analogy         & 200   & 0.1     & 0.01        & {-}      & 1-R        \\
        & RESCAL          & 200   & 0.1     & 0.01        & {-}      & 1-R        \\ \hline
 FB-237 & DistMult        & 200   & 0.01    & 0.01        & {-}      & 2          \\
        & TransE$_{l_1}$  & 200   & 0.1     & {-}         & 4.0      & 1          \\
        & ComplEx         & 150   & 0.1     & 0.001       & {-}      & 2          \\
        & Analogy         & 200   & 0.1     & 0.01        & {-}      & 2          \\
        & RESCAL          & 200   & 0.1     & 0.1         & {-}      & 1          \\  \hline
   WN18 & DistMult        & 200   & 0.1     & 0.01        & {-}      & 1          \\
        & TransE$_{l_1}$  & 150   & 0.1     & {-}         & 2.0      & 1-R        \\
        & ComplEx         & 200   & 0.1     & 0.01        & {-}      & 1-R        \\
        & Analogy         & 150   & 0.1     & 0.01        & {-}      & 1-R        \\
        & RESCAL          & 150   & 0.1     & 0.01        & {-}      & 1          \\ \hline
   WNRR & DistMult        & 100   & 0.1     & 0.001       & {-}      & 1          \\
        & TransE$_{l_1}$  & 100   & 0.1     & {-}         & 4.0      & 1-R        \\
        & ComplEx         & 200   & 0.1     & 0.001       & {-}      & 1          \\
        & Analogy         & 200   & 0.1     & 0.001       & {-}      & 1          \\
        & RESCAL          & 150   & 0.1     & 0.001       & {-}      & 1-R        \\ \hline
  \end{tabular}
  \caption{Hyperparameters for best models in PR}
  \label{tab:hyper-new}
\end{table}

\end{document}